\documentclass[twoside,11pt]{article}

\usepackage{jmlr2e}

\usepackage{multirow}
\usepackage{booktabs}
\usepackage{threeparttable}
\usepackage{pifont}  
\usepackage[table]{xcolor}
\usepackage{amssymb}
\usepackage{graphicx}
\usepackage{url}
\usepackage{caption}

\begin{document}

\begingroup
\renewcommand{\thefootnote}{\dag}
\footnotetext{These authors contributed equally to this work.}
\endgroup

\title{Awesome-OL: An Extensible Toolkit for Online Learning}

\author{%
  \name Zeyi Liu\textsuperscript{\dag} \email liuzy21@mails.tsinghua.edu.cn \\
  \addr Department of Automation, Tsinghua University\\
  Beijing, CHINA
  \AND
  \name Songqiao Hu\textsuperscript{\dag} \email hsq23@mails.tsinghua.edu.cn \\
  \addr Department of Automation, Tsinghua University\\
  Beijing, CHINA
  \AND
  \name Pengyu Han\textsuperscript{\dag} \email hpy24@mails.tsinghua.edu.cn \\
  \addr Department of Automation, Tsinghua University\\
  Beijing, CHINA
  \AND
  \name Jiaming Liu\textsuperscript{\dag} \email 23371007@buaa.edu.cn \\
  \addr Department of Computer Science, Beihang University\\
  Beijing, CHINA
  \AND
  \name Xiao He \email hexiao@tsinghua.edu.cn \\
  \addr Department of Automation, Tsinghua University\\
  Beijing, CHINA
}

\editor{}

\maketitle

\begin{abstract}
In recent years, online learning has attracted increasing attention due to its adaptive capability to process streaming and non-stationary data. To facilitate algorithm development and practical deployment in this area, we introduce Awesome-OL, an extensible Python toolkit tailored for online learning research. Awesome-OL integrates state-of-the-art algorithm, which provides a unified framework for reproducible comparisons, curated benchmark datasets, and multi-modal visualization. Built upon the scikit-multiflow open-source infrastructure, Awesome-OL emphasizes user-friendly interactions without compromising research flexibility or extensibility. The source code is publicly available at: \url{https://github.com/liuzy0708/Awesome-OL}.
\end{abstract}

\begin{keywords}
Stream learning, Non-stationary, Concept drift, Python
\end{keywords}

\section{Introduction}


The increasing complexity and non-stationarity of data streams, particularly in industrial predictive maintenance applications, have underscored the critical importance of developing advanced online learning methodologies \cite{hoi2021online,lu2018learning}. Compared to traditional batch-based approaches, online learning provides significant advantages by continuously adapting to evolving data distributions, enabling timely detection of anomalies, faults, and performance degradation, which are crucial for effective real-time decision-making \cite{liu2024review,zhou2022open,11007216}.

While numerous open-source toolkits have substantially facilitated machine learning model development and deployment by offering foundational frameworks and conventional algorithms, they generally fall short of fully addressing the specialized requirements of contemporary online learning scenarios, highlighting an urgent need for toolkits that explicitly support advanced online learning capabilities. Scikit-learn \cite{scikit-learn} is arguably the most widely used Python-based machine learning library, providing extensive algorithms for classification, regression, and clustering. Nevertheless, it primarily focuses on batch learning paradigms and lacks inherent mechanisms to handle streaming data and concept drift, both of which are critical in real-time, non-stationary environments.

In contrast,  MOA \cite{MOA} is a prominent Java-based platform specifically designed for data stream mining, encompassing a wide array of algorithms such as classification, regression, clustering, outlier detection, and drift detection. Despite its broad functionality and significant impact in academia and industry, MOA's Java-based implementation introduces interoperability constraints for the prevalent Python-driven machine learning research community. Furthermore, MOA does not yet integrate many recent algorithmic developments from the fast-evolving online learning literature. Another related toolkit, MEKA \cite{MEKA}, extends WEKA to facilitate multi-label learning scenarios, addressing the problem of simultaneously predicting multiple target variables per input instance. Nonetheless, MEKA is narrowly specialized in multi-label classification and does not aim to provide an inclusive platform suitable for broader online learning tasks.

After that, scikit-multiflow built a Python-based toolkit for handling data flow and concept drift, which was creatively combined Scikit-learn, MOA and MEKA advantages. However, the updates for scikit-multiflow were last made in 2018, which implies that it does not possess the current state-of-the-art models, but only includes some of the most classic online learning algorithms. River \cite{River} is a toolkit released in 2021 that is characterized by the high flexibility and ease of use of the framework, and its biggest feature is that it optimizes time complexity. It significantly reduces the time expenditure required for training the model. However, its limitations are still reflected in the fact that the model is not new enough.

In light of these limitations, we introduce Awesome-OL, a comprehensive and extensible Python toolkit tailored explicitly to online learning research. Awesome-OL integrates a diverse collection of recently proposed algorithms, systematically selected from the forefront of online learning research. This distinguishes Awesome-OL from existing toolkits, enabling researchers to conveniently access state-of-the-art methods for handling concept drift, label noise, semi-supervised streaming scenarios, and dynamic querying strategies. To enable researchers to use the toolkit more conveniently, we encapsulate all internal calls and provide users with four Jupyter Notebook Demos. In this way, users can not only shield the complex internal call relationships, but also make good use of the interaction of Jupyter Notebook to customize their required models and parameters.

\begin{table}[htbp]
    \centering
    \renewcommand{\arraystretch}{1.05} 
\setlength{\tabcolsep}{4pt}
\centering
\scriptsize
\caption{\textbf{\protect\makebox[\linewidth][c]{Overview of Representative Methodologies Implemented in Awesome-OL}}}

\label{table:methods}
\begin{tabular}{|l|l|c|c|c|c|c|c|>{\columncolor[gray]{0.92}}c>{\columncolor[gray]{0.92}}c|l|}
    \rowcolor[gray]{0.9}
    \hline
    \textbf{Year} & \textbf{Method} &
    \rotatebox{90}{Classification} &
    \rotatebox{90}{Strategy} &
    \rotatebox{90}{Regression} &
    \rotatebox{90}{Binary} &
    \rotatebox{90}{Multi-class} &
    \rotatebox{90}{Drift Adapt.} &
    \rotatebox{90}{scikit-multiflow} &
    \rotatebox{90}{Awesome-OL} &
    \textbf{Reference} \\
    \hline
    2025 & QRBLS          & \checkmark &        &        & \checkmark & \checkmark &        &        & \checkmark & \textit{\cite{QRBLS}} \\
    2024 & DMI-DD         &           & \checkmark &        & \checkmark & \checkmark & \checkmark &        & \checkmark & \textit{\cite{DMI-DD}} \\
    2024 & BLS-W          & \checkmark &        &        & \checkmark & \checkmark & \checkmark &        & \checkmark & \textit{\cite{DMI-DD}} \\
    2024 & DSA-AI         &           & \checkmark &        & \checkmark & \checkmark & \checkmark &        & \checkmark & \textit{\cite{DSA-AI}} \\
    2024 & DES            & \checkmark &        &        & \checkmark &            & \checkmark &        & \checkmark & \textit{\cite{DES}} \\
    2023 & IWDA           & \checkmark &        &        & \checkmark & \checkmark & \checkmark &        & \checkmark & \textit{\cite{IWDA}} \\
    2023 & MTSGQS         &           & \checkmark &        & \checkmark & \checkmark & \checkmark &        & \checkmark & \textit{\cite{MTSGQS}} \\
    2023 & CogDQS         &           & \checkmark &        & \checkmark & \checkmark & \checkmark &        & \checkmark & \textit{\cite{CogDQS}} \\
    2023 & OLI2DS         & \checkmark &        &        & \checkmark &            & \checkmark &        & \checkmark & \textit{\cite{OLI2DS}} \\
    2022 & ROALE-DI       & \checkmark & \checkmark &        & \checkmark & \checkmark & \checkmark &        & \checkmark & \textit{\cite{ROALE}} \\
    2021 & OSSBLS         & \checkmark &        &        & \checkmark & \checkmark & \checkmark &        & \checkmark & \textit{\cite{OSSBLS}} \\
    2021 & ISSBLS         & \checkmark &        &        & \checkmark & \checkmark & \checkmark &        & \checkmark & \textit{\cite{OSSBLS}} \\
    2020 & ACDWM          & \checkmark &        &        & \checkmark &            & \checkmark & \checkmark & \checkmark & \textit{\cite{ACDWM}} \\
    2019 & SRP            & \checkmark &        &        & \checkmark & \checkmark & \checkmark &        & \checkmark & \textit{\cite{SRP}} \\
    2019 & ARF            & \checkmark &        &        & \checkmark & \checkmark & \checkmark & \checkmark & \checkmark & \textit{\cite{gomes2017adaptive}} \\
    2019 & OALE           & \checkmark & \checkmark &        & \checkmark & \checkmark & \checkmark &        & \checkmark & \textit{\cite{OALE}} \\
    2006 & KNN            &           &        & \checkmark &        &        &        & \checkmark & \checkmark & \textit{\cite{bishop2006pattern}} \\
    2006 & MLP            & \checkmark &        &        & \checkmark & \checkmark & \checkmark & \checkmark & \checkmark & \textit{\cite{bishop2006pattern}} \\
    2001 & Hoeffding Tree &           &        & \checkmark &        &        &        & \checkmark & \checkmark & \textit{\cite{HoeffdingTree}} \\
    \hline
\end{tabular}

\end{table}

\section{Architecture}

\subsection{Classifier}

From the structure of classifiers, it mainly contains two functions: learning function and prediction function.

The first is the learning function: the \textbf{fit} function realizes the initial training of the model, generally using the gradient descent method. After the pre-training model is formed, the external can call the partial fit function to realize the online update function. This function will continue to update the model on the basis of the last updated model, which is also the core part of online learning. This part usually uses online gradient descent or online mirror descent to update the model parameters.

The second is the prediction function: in order to evaluate the effect of online learning, we need to know the prediction results of the model for each data. This requires us to call the predict function, which is used to predict the class of samples. In the predict function, the \textbf{predict\_proba} function is often called again, which will return the probability estimate of each class of samples.

After a round, the model will be updated, and the real tags and predicted results will be added to their respective lists and transferred to the visualization function as parameters after the training.

\subsection{Call relationship}

The call relationship is mainly concentrated in user, Online Learning demo, utils, classify model and visual tool.

In jupyter notebook, the user first selects the super parameters of the model, such as the number of samples, the number of pre-training, and the number of training rounds. Then, the user selects the framework, classify model, and strategy of online learning. After that, the program will automatically create an instance of the execution module, and then execute \textbf{experiment.run()} to enter the execution module.

In the execution module, first receive the super parameters passed in by the user, and then specify the path to save the output results. After the preparations are completed, call utils.py's \textbf{get\_clf, get\_str, and get\_stream} to obtain the classify models, strategies, and data streams respectively. Then enter the training phase, call fit, and then call predict and \textbf{partial\_fit} circularly. At the same time, record the prediction results.

\begin{figure}[htbp]
    \centering
    \includegraphics[width=0.90\linewidth]{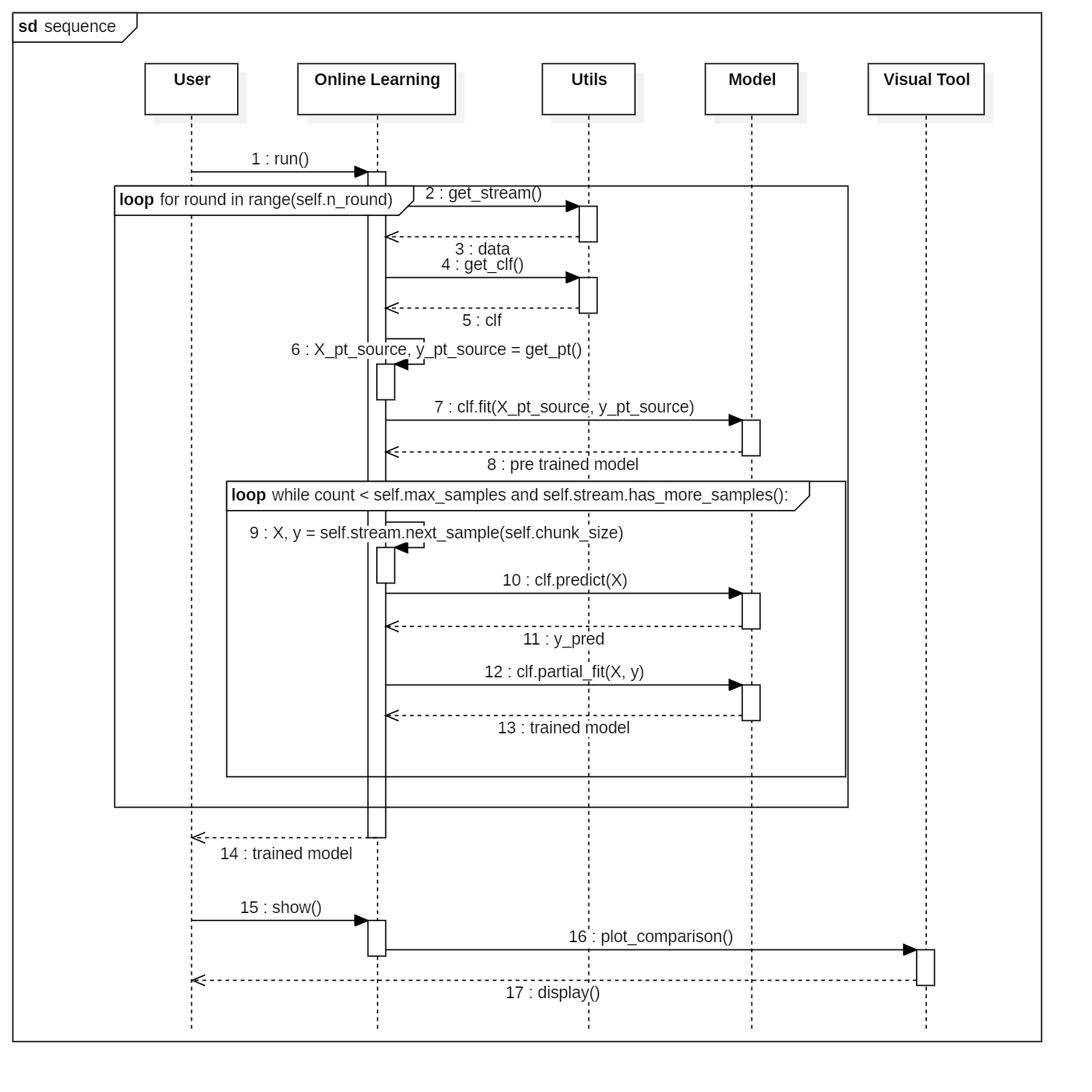}
    \caption{Sequence Diagram of the Execution Flow in Awesome-OL}
    \label{fig:data-flow}
\end{figure}

After the training, jupyter notebook will execute \textbf{experiment.show()} according to the user's choice, which will jump to \textbf{plot\_comparison.py} to generate visualization results. According to the prediction results, all the selected models will be drawn into icons and put together for comparison, and finally the visual results will be saved. The sequence of model training and evaluation in the Awesome-OL toolkit, including data streaming, online updates, and visualization, is illustrated in Figure~\ref{fig:data-flow}.

\section{Development}

Awesome-OL is released under the GNU General Public License (GPL), and users must comply with the terms of this open-source license.

\begin{itemize}
    \item All source code is publicly available on GitHub: \url{https://github.com/liuzy0708/Awesome-OL}. Please refer to the README for usage instructions and licensing terms.
    \item A dedicated website introduces key features, supported models, and environment setup tutorials. Visit: \url{https://thufdd.github.io}.
\end{itemize}


\bibliography{reference_info}

\end{document}